# Trajectory Tracking Using Frenet Coordinates with Deep Deterministic Policy Gradient

Tongzhou Jiang [1,*]
Independent Researcher
tojiang0111@gmail.com

Junyue Jiang[2]
University of Illinois at Urbana Champaign
USA
junyuej2@illinois.edu

Yuhui Jin[4]
Independent Researcher
yuhuijin4867@gmail.com

Lipeng Liu[2]
Peking University
Beijing, China
lipeng.liu@pku.edu.cn

Tianyao Zheng[3]
Independent Researcher
tian971227@gmail.com

Kunpeng Xu[5]
University of Sherbrooke
Canada
kunpeng.xu@usherbrooke.ca

*Abstract*—This paper studies the application of the DDPG algorithm in trajectory-tracking tasks and proposes a trajectory-tracking control method combined with Frenet coordinate system. By converting the vehicle's position and velocity information from the Cartesian coordinate system to Frenet coordinate system, this method can more accurately describe the vehicle's deviation and travel distance relative to the center line of the road. The DDPG algorithm adopts the Actor-Critic framework, uses deep neural networks for strategy and value evaluation, and combines the experience replay mechanism and target network to improve the algorithm's stability and data utilization efficiency. Experimental results show that the DDPG algorithm based on Frenet coordinate system performs well in trajectory-tracking tasks in complex environments, achieves high-precision and stable path tracking, and demonstrates its application potential in autonomous driving and intelligent transportation systems.

## I. INTRODUCTION

Trajectory tracking is a fundamental challenge in autonomous driving and robotics [1-3]. Traditional control methods, such as Pure Pursuit and Stanley methods, perform well in simple, predictable environments, but often perform poorly in complex dynamic situations [4-6]. These methods are model-based, which limits their adaptability and robustness in practical applications [7][8].

Reinforcement learning (RL) has emerged as an innovative solution to this problem, especially because it can learn and adjust control policies from environmental feedback [9-11]. One of the most promising methods in this field is deep reinforcement learning (DRL), which combines the advantages of deep learning with RL techniques [12][13]. One of the most effective DRL algorithms for continuous action space problems is deep deterministic policy gradient (DDPG), which combines elements of value-based learning (e.g., deep Q-networks) and policy gradient methods, enabling it to directly learn optimal policies in complex environments[14-18].

Reference [19] achieved accurate non-invasive diagnosis by applying convolutional neural networks (CNN) to snoring sound data, providing a more convenient solution for detecting nighttime breathing problems. Reference [20] uses a convolutional neural network to accurately identify various pills, and its practical application in daily life makes it a significant contribution to assistive technology for the visually impaired.

Reference [21] proposes an effective method for estimating relative states in multi-agent systems using distance data. The block multi-convex optimization method improves scalability, making it ideal for large-scale networks, especially in robots and self-driving cars. Reference [22] introduces a method for autonomous flight in cluttered environments by learning to dynamically adjust speed. This method focuses on improving navigation efficiency and safety and makes a valuable contribution to the field of autonomous aircraft.

In trajectory tracking, Frenet coordinates are often used because they simplify the control problem by transforming global Cartesian coordinates into a local reference frame relative to the trajectory [23][24]. This makes it easier to separate lateral and longitudinal control, so that each aspect of the vehicle's motion can be optimized independently [25][26]. By converting the vehicle's position and velocity into a Frenet framework, trajectory tracking becomes more manageable [27-29]. When used in conjunction with DDPG, the use of Frenet coordinates can improve the efficiency and effectiveness of trajectory tracking. DDPG is good at processing continuous control inputs and ensuring smooth transitions in steering and speed adjustments, which is critical for maintaining vehicle stability and safety [30-34].

This paper combines DDPG with Frenet coordinates to provide a powerful and scalable solution for trajectory tracking in autonomous driving, which can achieve precise control in complex environments.

## II. PROPOSED METHOD

### A. Cartesian frame to Serret-Frenet frame

As shown in Fig.1, since the s-axis is parallel to the lane line and perpendicular to the d-axis, the Frenet coordinate system is more convenient for describing the relationship between vehicle motion and road. Compared with the Cartesian coordinate system, the Frenet coordinate system can more easily calculate the distance the vehicle deviates from the road's center line and the distance traveled along the lane line without considering the road's curvature. This coordinate system is widely used in path planning research for self-driving cars, robots, and drones.

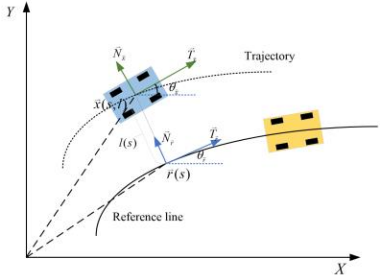

Fig 1. Frenet frame.

In the Frenet coordinate system: ($[s, \dot{s}, \ddot{s}, l, \dot{l}, \ddot{l}, l', l'']$), and Cartesian coordinate system: ($[x, y, v_x, a_x, \theta_x, k_x]$), the definitions are as follows:

$$\begin{cases} s = s_r \\ \dot{s} = \dfrac{v_x}{\cos(\theta_x - \theta_r)} \\ \ddot{s} = \dfrac{a_x \cos(\theta_x - \theta_r) - \dot{s}^2 \left[\left(k_x \dfrac{1-k_r l}{\cos(\theta_x - \theta_r)} - k_r\right) - (k_r' l + k_r l')\right]}{1 - k_r l} \\ l = \mathrm{sign}((y - y_r)\cos(\theta_r) - (x - x_r)\sin(\theta_r)) \\ \quad \cdot \sqrt{(x - x_r)^2 + (y - y_r)^2} \\ \dot{l} = v_x \sin(\theta_x - \theta_r) \\ \ddot{l} = a_x \sin(\theta_x - \theta_r) \\ l' = \dfrac{(1-k_r l)\tan(\theta_x - \theta_r)}{\sqrt{(x-x_r)^2 + (y-y_r)^2}} \\ l'' = -(k_r' l + k_r l')\tan(\theta_x - \theta_r) + \dfrac{(1-k_r l)}{\cos^2(\theta_x - \theta_r)}(k_x - k_r) \end{cases} \quad (1)$$

$s, \dot{s}, \ddot{s}$ : Frenet longitudinal coordinate, velocity, acceleration;

$l, \dot{l}, \ddot{l}$ : Frenet lateral coordinate, velocity, acceleration;

$l', l''$ : Derivative of $l$ to $s$, second derivative of $l$ to $s$;

$s_r$ : Arc length position along the reference path;

$\theta_x$ : Heading in the Cartesian coordinate system;

$\theta_r$ : Heading of the reference path at the position $s_r$;

$k_r$ : Curvature of the reference path at the position $s_r$;

$k_r'$ : Derivative of the curvature of the reference path at the position $s_r$;

$v_x$ : Linear velocity in the Cartesian coordinate system;

$a_x$ : Acceleration in the Cartesian coordinate system;

$x_r, y_r$ : Coordinates of the reference path at the position $s_r$.

### B. DDPG algorithm

The working principle of the DDPG algorithm involves the Actor network determining the optimal action for the current state, while the Critic network evaluates the value of the action chosen by the Actor network. Initially, the weights of the Actor and Critic networks are initialized, and their corresponding target networks are created. The target networks are copies of the main networks' weights but are updated more slowly to improve algorithm stability. During interaction with the environment, the algorithm obtains the state $s_t$, calculates the action $a_t$ through the Actor network, and, after executing the action, obtains the next state $s_{t+1}$ and reward $r_t$. The interaction tuple $s_t$, $a_t$, $r_t$, $s_{t+1}$ is stored in the experience replay buffer, from which small batches of samples are randomly drawn for training, computing target Q-values and updating the Critic network.

The loss function for the Critic network is based on the Bellman equation. The target Q-value is calculated as:

$$y_i = r_i + \gamma Q'(s_{i+1}, \mu'(s_{i+1} | \theta^{\mu'}) | \theta^{Q'}) \quad (2)$$

where $r_i$ is the reward received after taking action $a_i$ in state $s_i$, $\gamma$ is the discount factor, $Q'$ and $\mu'$ are the target networks for the Critic and Actor, respectively, and $\theta^{\mu'}$ and $\theta^{Q'}$ are the parameters of the Actor and Critic target networks. The loss function for the Critic network is:

$$L(\theta^Q) = \dfrac{1}{N} \sum_i \left( y_i - Q(s_i, a_i | \theta^Q) \right)^2 \quad (3)$$

The Actor network is updated by maximizing the Q-value output by the Critic network. The policy gradient is computed as:

$$\nabla_{\theta^\mu} J \approx \dfrac{1}{N} \sum_i \nabla_a Q(s, a | \theta^Q) \big|_{s=s_i, a=\mu(s_i)} \nabla_{\theta^\mu} \mu(s | \theta^\mu) \big|_{s=s_i} \quad (4)$$

where $J$ is the objective function of the policy, and $\theta^\mu$ are the parameters of the Actor network. The soft update of the target networks is performed as follows:

$$\begin{aligned} \theta^{Q'} &\leftarrow \tau \theta^Q + (1-\tau)\theta^{Q'} \\ \theta^{\mu'} &\leftarrow \tau \theta^\mu + (1-\tau)\theta^{\mu'} \end{aligned} \quad (5)$$

where $\tau$ is the soft update coefficient, typically set to a small value.

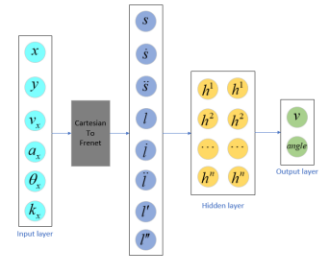

Fig 2. Input and output structure.

By combining deep learning and reinforcement learning techniques, the DDPG algorithm excels in continuous action spaces. Its Actor-Critic framework and experience replay mechanism ensure algorithm stability and data utilization efficiency. In various complex control tasks, the DDPG algorithm can effectively learn optimal policies, making it widely applicable in fields such as robotic control and autonomous driving. With the powerful expressive capabilities of deep neural networks, the DDPG algorithm can effectively explore and learn in high-dimensional state and action spaces, achieving optimal control for complex tasks.

*C. Loss Function*

Our goal is to reduce the lateral error between the vehicle's current position and the reference trajectory, and to minimize the lateral velocity and acceleration to ensure that the vehicle does not deviate during driving. So the loss function is:

$$L = l + \dot{l} + \ddot{l} \qquad (6)$$

We transform the vehicle's position from the Cartesian coordinate system to the Frenet coordinate system, and re-express the vehicle's position and speed information in terms of lateral and longitudinal position and speed information as the environmental input information of the DDPG algorithm, thereby replacing the original loss function calculated based on the Cartesian coordinate system. In the process of backpropagation of the neural network, the correct gradient direction can be found, making the algorithm converge faster.

## III. EXPERIMENTS

To verify our trajectory tracking algorithm, we conducted simulation experiments in the Gazebo simulation environment. Gazebo is a powerful simulation platform that can realistically simulate physical and sensor data in the real world. We built a virtual environment with roads and vehicles in Gazebo.

In our experiments, we used the SGD optimizer to update the model parameters and trained with a batch size of 128. We trained several other comparison algorithms on the same dataset to evaluate the effect of the proposed DDPG algorithm. In order to comprehensively compare the performance of these algorithms, we recorded their average losses for quantitative analysis. Through this detailed recording and comparison, we were able to deeply analyze the performance of each algorithm in different scenarios. This not only helped us verify the effectiveness of the DDPG algorithm, but also provided valuable reference information for subsequent optimization.

Secondly, we selected a representative trajectory for qualitative analysis. We show the effect of trajectory tracking by different algorithms and the position deviation of the algorithm relative to the original trajectory in the figure.

Our evaluation method emphasizes systematicity and comprehensiveness, ensuring that we can identify and understand the advantages and disadvantages of each algorithm, thereby promoting further improvement and development of autonomous driving path tracking control algorithms.

*A. Quantitative Analysis*

We used two common reinforcement learning algorithms, PPO and DDPG, respectively, and conducted experiments in their original form and in the form of adding the Frenet coordinate system. The Frenet coordinate system is designed to improve the accuracy of trajectory tracking, thereby improving the performance of the algorithm.

The trend of the loss value during training is shown in Fig. 3, which shows the change of the loss value of each method with the number of iterations during training. The loss values of all methods are decreasing with the increase of the number of iterations, indicating that each method is constantly learning and optimizing. The loss value decreases rapidly in the early stage of training and tends to be stable in the later stage. In particular, the DDPG with the introduction of Frenet has always maintained the lowest loss value throughout the training process, further verifying its superiority.

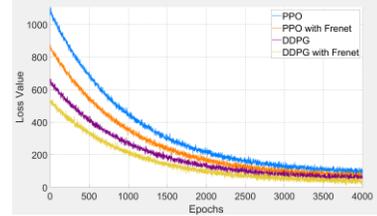

Fig 3. Loss value during training process

*B. Qualitative analysis*

In this experiment, we used four different trajectory tracking algorithms for comparison. During the experiment, the vehicle was tracked along the predefined reference path, and the actual path and lateral error of the trajectory are shown in Fig. 4(a) and Fig. 4(b), respectively.

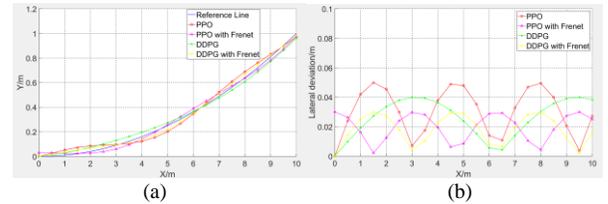

Fig 4. (a) Tracking performance of different algorithms; (b) Lateral error of tracking with different algorithms

Fig. 4(a) shows the performance of different algorithms when tracking the predefined reference path. As can be seen from the Fig.4, the DDPG algorithm based on Frenet coordinate system (yellow curve) can track the reference path more accurately and stay near the reference path. Its tracking accuracy is significantly better than the other three algorithms. This shows that the DDPG algorithm based on Frenet coordinate system has higher accuracy and stability when dealing with complex path tracking tasks.

Fig. 4(b) shows the lateral errors generated by different algorithms during the tracking process. The results show that the DDPG algorithm based on Frenet coordinate system has the smallest lateral error, smooth curve, and small fluctuation, which means that the algorithm can not only effectively reduce the tracking error, but also provide smoother trajectory tracking performance and improve the overall tracking effect.

In summary, by comparing the trajectory tracking performance of different algorithms under the same conditions, this study shows that the DDPG algorithm based on the Frenet coordinate system performs well in trajectory tracking tasks, and can track predefined paths with high accuracy and stability, while significantly reducing lateral errors, providing strong support for path planning and control of autonomous vehicles.

IV. CONCLUSIONS

This paper proposes an innovative path tracking controller and designs a steering method that combines DDPG and Frenet coordinate systems to adapt to various complex tracking scenarios. During the tracking process, the method also comprehensively considers the vehicle's speed and position information, allowing the steering control to dynamically adapt to the vehicle's current speed. Experiments conducted on a driving simulator verify the effectiveness and superior performance of the controller.